\documentclass{article}

\usepackage[utf8]{inputenc} 
\usepackage[T1]{fontenc}    
\usepackage{lmodern}
\usepackage{url}            
\usepackage{booktabs}       
\usepackage{amsfonts}       
\usepackage{nicefrac}       
\usepackage{microtype}      
\usepackage{setspace}
\usepackage{caption}
\usepackage{enumitem}
\usepackage{xfrac}
\usepackage{amsmath}
\usepackage[square,semicolon]{natbib}
\usepackage[backref=page]{hyperref}
\usepackage{xcolor}
\usepackage{rotating}
\usepackage[paper=8.5in:11in]{typearea}
\usepackage[margin=1.0in]{geometry}
\usepackage{amssymb}
\usepackage{comment}

\usepackage{multicol}
\usepackage{microtype}
\usepackage{graphicx}
\usepackage[procnames]{listings}
\definecolor{keywords}{RGB}{255,0,90}
\definecolor{comments}{RGB}{0,0,113}
\definecolor{red}{RGB}{160,0,0}
\definecolor{green}{RGB}{0,100,0}
 
\definecolor{darkblue}{rgb}{0, 0.2, 0.7}
\hypersetup{
    colorlinks = true,
    linkcolor = darkblue,
    anchorcolor = darkblue,
    citecolor = darkblue,
    filecolor = darkblue,
    urlcolor = darkblue
}

\title{\vspace{-2em}%
  \hrule height 4pt%
  \vskip 0.25in%
  \vskip -\parskip%
  \textbf{
  Talking-Heads Attention
  }%
  \vskip 0.2in%
  \vskip -\parskip%
  \hrule height 1pt%
  \vskip 0.09in}

\author{
Noam Shazeer\thanks{Noam Shazeer devised the talking-heads architecture, ran the T5 experiments and wrote most of the paper. Zhenzhong Lan had the initial idea of talking-heads attention, designed and coordinated part of the experiments. Youlong Cheng reproduced BERT in MeshTensorFlow and run all the talking heads experiments for MeshTensorFlow BERT. Nan Ding ran the ALBERT experiments. Le Hou visualized and analyzed the learned weights of talking-heads.}, \\
    Google \\
    \tt{noam@google.com}  \\ \and
    Zhenzhong Lan\footnotemark[1] \\ 
    Google \\ 
    \tt{lanzhzh@google.com} \\ \and
    Youlong Cheng\footnotemark[1] \\ 
    Google \\ 
    \tt{ylc@google.com} \\ \and
    Nan Ding\footnotemark[1] \\ 
    Google \\ 
    \tt{dingnan@google.com} \\ \and
    Le Hou\footnotemark[1] \\
    Google \\
    \tt{lehou@google.com}}

\begin{document}

\maketitle

\begin{abstract}

We introduce "talking-heads attention" - a variation on multi-head attention which includes linear projections across the attention-heads dimension, immediately before and after the softmax operation.   While inserting only a small number of additional parameters and a moderate amount of additional computation, talking-heads attention leads to better perplexities on masked language modeling tasks, as well as better quality when transfer-learning to language comprehension and question answering tasks.
\end{abstract}


\section{Introduction}
Neural Attention was introduced by \citep{1409.0473} as a way of extracting information from variable-length representations.  The Transformer model \citep{Vas17} uses "multi-head" attention, consisting of multiple attention layers ("heads") in parallel, each with different projections on its inputs and outputs.  By using a dimensionality reduction in the input projections, the computational cost is kept similar to that of basic attention.  Quality is improved, presumably due to the ability to attend to multiple positions simultaneously based on multiple different types of relationships. 

As noted in \citep{Vas17}\footnote{Section (A) of table 3 in \citep{Vas17}.  Also the first sections of tables \ref{tab:t5} and \ref{tab:albert} of this paper.}, taking this process to the extreme (more attention heads projected to lower dimensionality) becomes counterproductive.  We believe that this is due to the fact that the query-vectors and key-vectors become so low-dimensional that their dot product can no longer constitute an informative matching function. 


In this paper, we introduce a new variant, "talking-heads attention", that addresses this problem by inserting a learned linear projection across the attention-heads dimension of the attention-logits tensor.  This allows each attention function to depend on all of the keys and queries.  We also insert a second such projection immediately following the softmax.

We show experimentally that inserting these "talking-heads" projections leads to better perplexities on masked language modeling tasks, as well as better quality when transfer-learning to language comprehension and question answering tasks.



\section{Notation}
In our pseudocode, we use capital letters to represent tensors and lower-case letters to represent their dimensions.  Each tensor is followed by a dimension list in brackets.  For example, a 4-dimensional image-tensor with (batch, height, width, channels) dimensions would be written as:
\begin{lstlisting}[language=Python]
  X[b, h, w, c]
\end{lstlisting}

We use \textbf{einsum} notation for generalized contractions between tensors of arbitrary dimension.  The computation is numerically equivalent to broadcasting each input to have the union of all dimensions, multiplying component-wise, and summing across all dimensions not in the output.  Rather than identifying the dimensions by an equation, as in TensorFlow and numpy, the dimensions are indentified by the dimension-list annotations on the arguments and on the result.  For example, multiplying two matrices would be expressed as: \\
\begin{lstlisting}[language=Python]
  Y[a, c] = einsum(X[a, b], W[b, c])
\end{lstlisting}

\section{Review of Attention Algorithms}

\subsection{Dot-Product Attention}

Simple dot-product attention can be described by the pseudocode below.  The logits L are computed as the dot-products of the query-vectors and the memory-vectors.  For each query, the logits are passed through a softmax function to produce weights, and the different memory-vectors are averaged together, weighted by those weights.   In this code, we show the case where there are $n$ different queries all attending to the same $m$ memory-vectors.  If there is only one query, the code is identical except that the "n" dimension is removed from all tensors.  \\
\begin{minipage}{\textwidth}
\begin{lstlisting}[language=Python]
def DotProductAttention(
    X[n, d],   # n query-vectors with dimensionality d
    M[m, d]):  # m memory-vectors with dimensionality d
  L[n, m] = einsum(X[n, d], M[m, d])           # Attention logits
  W[n, m] = softmax(L[n, m], reduced_dim=m)    # Attention weights
  Y[n, d] = einsum(W[n, m], M[m, d])
  return Y[n, d]
\end{lstlisting}
\end{minipage}

\subsection{Dot-Product Attention With Projections}

\citep{Vas17} propose a dimensionality-reduction to reduce the computational complexity of the attention algorithm.  In this version, instead of computing the attention algorithm directly on the inputs $X$ and $M$, we first project the inputs using the learned linear projections $P_q$, $P_k$ and $P_v$, to produce lower-dimensional query-vectors, key-vectors and value-vectors $Q$, $K$ and $V$.  We use a fourth learned linear projection, $P_o$, to produce the output. \\

\begin{minipage}{\linewidth}
\begin{lstlisting}[language=Python]
def DotProductAttentionWithProjections(
    X[n, d_X],      # n vectors with dimensionality d_X
    M[m, d_M],      # m vectors with dimensionality d_M
    P_q[d_X, d_k],  # learned linear projection to produce queries
    P_k[d_M, d_k],  # learned linear projection to produce keys
    P_v[d_M, d_v],  # learned linear projection to produce values
    P_o[d_Y, d_v]): # learned linear projection of output
  Q[n, d_k] = einsum(X[n, d_X], P_q[d_X, d_k])  # queries
  K[m, d_k] = einsum(M[m, d_M], P_k[d_M, d_k])  # keys
  V[m, d_v] = einsum(M[m, d_M], P_v[d_M, d_v])  # values
  L[n, m] = einsum(Q[n, d_k], K[m, d_k])     # Attention logits
  W[n, m] = softmax(L[n, m], reduced_dim=m)  # Attention weights
  O[n, d_v] = einsum(W[n, m], V[m, d_v])
  Y[n, d_Y] = einsum(O[n, d_v], P_o[d_Y, d_v])
  return Y[n, d_Y]
\end{lstlisting}
\end{minipage}

\subsection{Multi-Head Attention}
The multi-head attention described in \citep{Vas17} consists of the sum of multiple parallel attention layers.  This can be represented by adding a "heads" dimension h to the above computation. \\
\begin{minipage}{\linewidth}
\begin{lstlisting}[language=Python]
def MultiHeadAttention(
    X[n, d_X],         # n vectors with dimensionality d_X
    M[m, d_M],         # m vectors with dimensionality d_M
    P_q[d_X, d_k, h],  # learned linear projection to produce queries
    P_k[d_M, d_k, h],  # learned linear projection to produce keys
    P_v[d_M, d_v, h],  # learned linear projection to produce values
    P_o[d_Y, d_v, h]): # learned linear projection of output
  Q[n, d_k, h] = einsum(X[n, d_X], P_q[d_X, d_k, h])  # queries   h*n*d_X*d_k
  K[m, d_k, h] = einsum(M[m, d_M], P_k[d_M, d_k, h])  # keys      h*m*d_M*d_k
  V[m, d_v, h] = einsum(M[m, d_M], P_v[d_M, d_v, h])  # values    h*m*d_M*d_v
  L[n, m, h] = einsum(Q[n, d_k, h], K[m, d_k, h])     # logits    h*n*m*d_k
  W[n, m, h] = softmax(L[n, m, h], reduced_dim=m)     # weights
  O[n, d_v, h] = einsum(W[n, m, h], V[m, d_v, h])     #           h*n*m*d_v
  Y[n, d_Y] = einsum(O[n, d_v, h], P_o[d_Y, d_v, h])  # output    h*n*d_Y*d_v
  return Y[n, d_Y]
\end{lstlisting}
\end{minipage}

The pseudo-code above illustrates the practical step-by-step computation of multi-head attention.  The costs of the einsum operations (the number of multiplications in a naive implementation) are shown in the comments.  The equivalent pseudo-code below uses multi-way einsums and is more concise: \\
\begin{minipage}{\linewidth}
\begin{lstlisting}[language=Python]
def MultiHeadAttentionConcise(X, M, P_q, P_k, P_v, P_o):
  L[n, m, h] = einsum(X[n, d_X], 
                      M[m, d_M],
                      P_q[d_X, d_k, h],
                      P_k[d_M, d_k, h])
  W[n, m, h] = softmax(L[n, m, h], reduced_dim=m)
  Y[n, d] = einsum(W[n, m, h],
                   M[m, d_M],
                   P_v[d_M, d_v, h],
                   P_o[d_Y, d_v, h])
  return Y[n, d_Y]
\end{lstlisting}
\end{minipage}

Note: \citep{Vas17} include a constant scaling factor on the logits.  We omit this in our code, as it can be folded into the linear projections $P_q$ or $P_k$.  

\section{Talking-Heads Attention}
\label{sec:th}
In multi-head attention, the different attention heads perform separate computations, which are then summed at the end.  Our new variation, which we call "Talking-Heads Attention" breaks that separation.  We insert two additional learned linear projections, ${P_l}$ and ${P_w}$, which transform the attention-logits and the attention-weights respectively, moving information across attention heads.  \footnote{Appendix \ref{sec:dp} presents a variation on this, where the projection matrices themselves are input-dependent.}  Instead of one "heads" dimension $h$ across the whole computation, we now have three separate heads dimensions: $h_k$, $h$, and $h_v$, which can optionally differ in size (number of "heads").  $h_k$ refers to the number of attention heads for the keys and the queries.  $h$ refers to the number of attention heads for the logits and the weights, and $h_v$ refers to the number of attention heads for the values.  The algorithm is shown by the pseudo-code below.  The costs of the einsum operations are shown in the comments. \\
\begin{minipage}{\linewidth}
\begin{lstlisting}[language=Python]
def TalkingHeadsAttention(
    X[n, d_X],           # n vectors with dimensionality d_X
    M[m, d_M],           # m vectors with dimensionality d_M
    P_q[d_X, d_k, h_k],  # learned linear projection to produce queries
    P_k[d_M, d_k, h_k],  # learned linear projection to produce keys
    P_v[d_M, d_v, h_v],  # learned linear projection to produce values
    P_o[d_Y, d_v, h_v],  # learned linear projection of output
    P_l[h_k, h],         # talking-heads projection for logits
    P_w[h, h_v]):        # talking-heads projection for weights
  Q[n, d_k, h_k] = einsum(X[n, d_X], P_q[d_X, d_k, h_k])  # queries        n*d_X*d_k*h_k
  K[m, d_k, h_k] = einsum(M[m, d_M], P_k[d_M, d_k, h_k])  # keys           m*d_M*d_k*h_k
  V[m, d_v, h_v] = einsum(M[m, d_M], P_v[d_M, d_v, h_v])  # values         m*d_M*d_v*h_v
  J[n, m, h_k] = einsum(Q[n, d_k, h_k], K[m, d_k, h_k])   # dot prod.      n*m*d_k*h_k
  L[n, m, h] = einsum(J[n, m, h_k], P_l[h_k, h])    # Talking-heads proj.  n*m*h*h_k
  W[n, m, h] = softmax(L[n, m, h], reduced_dim=m)   # Attention weights
  U[n, m, h_v] = einsum(W[n, m, h], P_w[h, h_v])    # Talking-heads proj.  n*m*h*h_v
  O[n, d_v, h_v] = einsum(U[n, m, h_v], V[m, d_v, h_v])   #                n*m*d_v*h_v
  Y[n, d_Y] = einsum(O[n, d_v, h_v], P_o[d_Y, d_v, h_v])  #                n*d_Y*d_v*h_v
  return Y[n, d_Y]
\end{lstlisting}
\end{minipage}
Again, we can write this more concisely using multi-way einsum operations: \\
\begin{minipage}{\linewidth}
\begin{lstlisting}[language=Python]
def TalkingHeadsAttentionConcise(X, M, P_q, P_k, P_v, P_o, P_l, P_w):
  L[n, m, h] = einsum(X[n, d_X],
                      M[m, d_M],
                      P_q[d_X, d_k, h_k],
                      P_k[d_M, d_k, h_k],
                      P_l[h_k, h])
  W[n, m, h] = softmax(L[n, m, h], reduced_dim=m)
  Y[n, d_Y] = einsum(W[n, m, h],
                     M[m, d_M],
                     P_v[d_M, d_v, h_v],
                     P_o[d_Y, d_v, h_v],
                     P_w[h, h_v])
  return Y[n, d_Y]
\end{lstlisting}
\end{minipage}

\section{Complexity Analysis}

If we assume that $d_X=d_Y$, then the number of scalar multiplications in multi-head attention is:
\[h\cdot(d_k + d_v)\cdot(n\cdot d_X + m \cdot d_M + n \cdot m)\]

The number of scalar multiplications in talking-heads attention is:
\[(d_k \cdot h_k + d_v \cdot h_v)\cdot(n\cdot d_X + m \cdot d_M + n \cdot m)  + n \cdot m \cdot h \cdot (h_k + h_v)\]
The first term in this expression matches up with the cost of multi-head attention.  The second term is due to the talking-heads projections.  If $h < d_k$ and $h < d_v$, the the costs of the new talking-heads projections, $n \cdot m \cdot h \cdot h_k$ and $n \cdot m \cdot h \cdot h_v$ are less than the existing terms $n \cdot m \cdot d_k \cdot h_k$ and $n \cdot m \cdot d_v \cdot h_v$, respectively.

In practice, the talking-heads projections may be expensive on some neural-network accelerators due to the small dimension sizes involved.

\section{One More Way To Look At It}
\label{sec:gbma}

Mathematically, one can view multi-head attention and talking-heads attention as two special cases of the same general function, which we will call "general bilinear multihead attention" (GBMA).  GBMA uses two three-dimensional parameter tensors, as defined in the pseudocode below.  Due to its high computational cost, GBMA may have no practical use.   Multi-head attention is mathematically equivalent to a version of GBMA where each of the two parameter tensors is expressed as the product of two factors, as shown below.  Talking-heads attention is mathematically equivalent to a version of GBMA where each of the two parameter tensors is expressed as the product of three factors, as shown below. \\
\begin{minipage}{\linewidth}
\begin{lstlisting}[language=Python]
def GeneralBilinearMultiheadAttention(
    X[n, d_X],   # n vectors with dimensionality d_X
    M[m, d_M],   # m vectors with dimensionality d_M
    P[d_X, d_M, h],  # learned parameters
    Q[d_M, d_Y, h]): # learned parameters
  L[n, m, h] = einsum(X[n, d_X],  M[m, d_M], P[d_X, d_M, h])
  W[n, m, h] = softmax(L[n, m, h], reduced_dim=m)
  Y[n, d_Y] = einsum(W[n, m, h], M[m, d_M], Q[d_M, d_Y, h])
  return Y[n, d_Y]

def MultiHeadAttentionInefficient(X, M, P_q, P_k, P_v, P_o):
  P[d_X, d_M, h] = einsum(P_q[d_X, d_k, h], P_k[d_M, d_k, h])
  Q[d_M, d_Y, h] = einsum(P_v[d_M, d_v, h], P_o[d_Y, d_v, h])
  return GeneralBilinearMultiheadAttention(X, M, P, Q)

def TalkingHeadsAttentionInefficient(X, M, P_q, P_k, P_v, P_o, P_l, P_w):
  P[d_X, d_M, h] = einsum(P_q[d_X, d_k, h_k], P_k[d_M, d_k, h_k], P_l[h_k, h])
  Q[d_M, d_Y, h] = einsum(P_v[d_M, d_v, h_v], P_o[d_Y, d_v, h_v], P_w[h, h_v])
  return GeneralBilinearMultiheadAttention(X, M, P, Q)
\end{lstlisting}
\end{minipage}

\section{Experiments}

\subsection{Text-to-Text Transfer Transformer (T5)}
\label{sec:t5}
We test various configurations of multi-head attention and talking-heads attention on the transfer-learning setup from \citep{raffel2019exploring}.  An encoder-decoder transformer model \citep{Vas17} is pre-trained on a denoising objective of predicting missing text segments (average span length 3) from the C4 dataset \citep{raffel2019exploring} \footnote{This is identical to one of the training objecives described in \citep{raffel2019exploring}}, and subsequently fine-tuned on various language understanding tasks.   We use the same code base and model architecture as the base model from \citep{raffel2019exploring}.  The encoder and decoder each consist of 12 layers, with $d_{model}=768$ and $d_{ff}=3072$.  Each encoder layer contains a multi-head self-attention layer, and each decoder layer contains a multi-head self-attention layer and a multi-head attention-over-encoder layer.  For their base model, \citep{raffel2019exploring} follow \citep{devlin2018bert} and others, using $h=12$ and $d_k = d_v = 64$ for all of these attention layers.  We compare this setting to a variety of other configurations of multi-head and talking-heads attention, as detailed in table \ref{tab:t5}.

Similar to \citep{raffel2019exploring}, we pre-train our models for 524288 steps.  Each training batch consists of 128 examples, each of which has an input of 512 tokens and an output of 114 tokens, the output containing multiple spans of tokens which were deleted from the input.   Similarly to \citep{raffel2019exploring}, we use the Adafactor optimizer \citep{shazeer2018adafactor} and an inverse-square-root learning-rate schedule.  We also decay the learning rate linearly for the final 10 percent of the training steps.   Our main departure from \citep{raffel2019exploring} is that we, as suggested by \citep{lan2019albert},  use no dropout during pre-training.  We find this to produce superior results. We compute the log-perplexity on the training objective on a held-out shard of C4, which we believe to be a good indicator of model quality.  For each configuration, we train one model for the "full" 524288 steps and four models for a shorter time (65536 steps) to measure inter-run variability.  The results are listed in table \ref{tab:t5}.

We then fine-tune each of the models on an examples-proportional mixture of SQUAD \citep{rajpurkar2016squad}, GLUE \citep{wang2018glue} and SuperGlue \citep{wang2019superglue}.  Fine-tuning consists of 131072 additional steps with a learning rate of $10^{-3}$.  Following \citep{raffel2019exploring}, we use a dropout rate $0.1$ on the layer outputs, feed-forward hidden-layers and attention weights.  The embedding matrix (also used as the projection in the final classifier layer) is fixed during fine-tuning.  Tables \ref{tab:t5}, \ref{tab:t5_2}, \ref{tab:t5_3} and \ref{tab:t5_4} include results for SQUAD and MNLI-m.  Results for all other tasks are listed in the appendix.

\subsubsection{Multi-Head vs Talking-Heads Attention}

\begin{table}[h]
\caption{Multi-Head vs. Talking-Heads attention on T5}
\label{tab:t5}
\begin{center}
\vspace{-2mm}
\scalebox{0.75}{
\begin{tabular}{cccccc|cc|cc|ccc}
\hline\rule{0pt}{2.0ex}
     &       &     &       &       &       & ln(PPL) & ln(PPL) & SQUAD  &  & step & parameters & multiplies \\
     &       &     &       &       &       & 65536   & 524288  & v1.1   & MNLI-m & time & per  & per att. layer\\
     & $h_k$ & $h$ & $h_v$ & $d_k$ & $d_v$ & steps   & steps   & dev-f1 & dev    & (s) & att. layer  & (n=m=512) \\
\hline
multi-head &  & 6 &  & 128 & 128 & 2.010 (0.005) & 1.695 & 89.88 & 85.34 & \textbf{0.14} & 2359296 & $1.611 \cdot 10^9$ \\
multi-head &  & 12 &  & 64 & 64 & 1.982 (0.003) & 1.678 & 90.87 & 86.20 & 0.15 & 2359296 & $1.611 \cdot 10^9$ \\
multi-head &  & 24 &  & 32 & 32 & 1.989 (0.009) & 1.669 & 91.04 & 86.41 & 0.17 & 2359296 & $1.611 \cdot 10^9$ \\
multi-head &  & 48 &  & 16 & 16 & 2.011 (0.004) & 1.682 & 90.35 & 85.32 & 0.21 & 2359296 & $1.611 \cdot 10^9$ \\
\hline
talking-heads & 6 & 6 & 6 & 128 & 128 & 1.965 (0.009) & 1.659 & 90.51 & 85.99 & 0.16 & 2359368 & $1.629 \cdot 10^9$ \\
talking-heads & 12 & 12 & 12 & 64 & 64 & 1.932 (0.004) & 1.641 & 91.38 & 86.19 & 0.18 & 2359584 & $1.686 \cdot 10^9$ \\
talking-heads & 24 & 24 & 24 & 32 & 32 & 1.910 (0.001) & 1.624 & 91.83 & 87.42 & 0.22 & 2360448 & $1.913 \cdot 10^9$ \\
talking-heads & 48 & 48 & 48 & 16 & 16 & \textbf{1.903} (0.006) & 1.603 & \textbf{91.90} & \textbf{87.50} & 0.32 & 2363904 & $2.819 \cdot 10^9$ \\
\hline
\hline
multi-head &  & 24 &  & 64 & 64 & 1.950 (0.005) & 1.625 & 91.46 & 86.58 & 0.22 & 4718592 & $3.221 \cdot 10^9$ \\
general bilinear &  & 12 &  & 768 & 768 & 1.921 (0.011) & \textbf{1.586} & 90.83 & 86.50 & 0.47 & 14155776 & $7.650 \cdot 10^9$ \\
\hline\hline
\citep{raffel2019exploring} &  & 12 &  & 64 & 64 &  &  & 89.66 & 84.85 &  &  2359296 & $1.611 \cdot 10^9$\\
\hline
\end{tabular}
}
\end{center}
\end{table}

In table \ref{tab:t5}, we compare multi-head attention to talking-heads attention.   For each of the two algorithms, we test versions with 6, 12, 24 and 48 heads.  Following \citep{Vas17}, as we increase the number of heads, we decrease the key/value dimensionality $d_k$ and $d_v$, so as to keep the number of parameters constant.  For each number of heads, talking-heads attention improves over multi-head attention on all quality metrics.

Additionally, multi-head attention gets worse as we increase the number of heads from 24 to 48 and decrease the key and value dimensionalty from 32 to 16, while talking-heads attention gets better.  We presume that this is due to the keys being too short to produce a good matching signal.

For additional comparison, we include in table \ref{tab:t5} two models with significantly more parameters and computation in the attention layers.  In the first, we double the number of heads in our baseline model from $12$ to $24$ without reducing $d_k$ and $d_v$, resulting in a multi-head attention layer with double the parameters and double the computation.  In the second, we use "general bilinear multihead attention", as described in section \ref{sec:gbma}.

We also list the results from \citep{raffel2019exploring}.  We believe that their results are worse due to their use of dropout during pre-training.

\subsubsection{Varying the Heads-Dimensions Separately}

\begin{table}[h]
\caption{Talking-heads attention has three "heads" dimensions that can vary independently.}
\label{tab:t5_2}
\begin{center}
\vspace{-2mm}
\scalebox{0.75}{
\begin{tabular}{cccccc|cc|cc|ccc}
\hline\rule{0pt}{2.0ex}
     &       &     &       &       &       & ln(PPL) & ln(PPL) & SQUAD  &  & step & parameters & multiplies \\
     &       &     &       &       &       & 65536   & 524288  & v1.1   & MNLI-m & time & per  & per att. layer\\
     & $h_k$ & $h$ & $h_v$ & $d_k$ & $d_v$ & steps   & steps   & dev-f1 & dev    & (s) & att. layer  & (n=m=512) \\
\hline
talking-heads & 6 & 6 & 6 & 128 & 128 & 1.965 (0.009) & 1.659 & 90.51 & 85.99 & 0.16 & 2359368 & $1.629 \cdot 10^9$ \\
talking-heads & 6 & 24 & 6 & 128 & 128 & 1.941 (0.009) & 1.641 & 90.91 & 86.29 & 0.18 & 2359584 & $1.686 \cdot 10^9$ \\
talking-heads & 24 & 6 & 24 & 32 & 32 & 1.959 (0.008) & 1.667 & 90.77 & 86.15 & 0.20 & 2359584 & $1.686 \cdot 10^9$ \\
talking-heads & 6 & 24 & 24 & 128 & 32 & 1.939 (0.011) & 1.633 & 91.06 & 86.31 & 0.20 & 2360016 & $1.799 \cdot 10^9$ \\
talking-heads & 24 & 24 & 6 & 32 & 128 & 1.931 (0.013) & 1.628 & 90.98 & 86.81 & 0.21 & 2360016 & $1.799 \cdot 10^9$ \\
talking-heads & 24 & 24 & 24 & 32 & 32 & \textbf{1.910} (0.001) & \textbf{1.624} & \textbf{91.83} & \textbf{87.42} & 0.22 & 2360448 & $1.913 \cdot 10^9$ \\
\hline
\end{tabular}
}
\end{center}
\end{table}

In table \ref{tab:t5_2}, we experiment with independently varying the sizes of the three heads-dimensions.  From the results, it appears that all three are good to increase, but that the softmax-heads dimension $h$ is particularly important.

\subsubsection{Logits-Projection Only and Weights-Projection Only}

\begin{table}[h]
\caption{The logits-projection and the weights-projection can be employed separately.}
\label{tab:t5_3}
\begin{center}
\vspace{-2mm}
\scalebox{0.75}{
\begin{tabular}{cccccc|cc|cc|ccc}
\hline\rule{0pt}{2.0ex}
     &       &     &       &       &       & ln(PPL) & ln(PPL) & SQUAD  &  & step & parameters & multiplies \\
     &       &     &       &       &       & 65536   & 524288  & v1.1   & MNLI-m & time & per  & per att. layer\\
     & $h_k$ & $h$ & $h_v$ & $d_k$ & $d_v$ & steps   & steps   & dev-f1 & dev    & (s) & att. layer  & (n=m=512) \\
\hline
multi-head &  & 24 &  & 32 & 32 & 1.989 (0.009) & 1.669 & 91.04 & 86.41 & 0.17 & 2359296 & $1.611 \cdot 10^9$ \\
project logits & 24 & 24 &  & 32 & 32 & 1.969 (0.004) & 1.652 & 91.29 & 85.86 & 0.23 & 2359872 & $1.762 \cdot 10^9$ \\
project weights &  & 24 & 24 & 32 & 32 & 1.951 (0.009) & 1.636 & 91.03 & 86.12 & 0.23 & 2359872 & $1.762 \cdot 10^9$ \\
talking-heads & 24 & 24 & 24 & 32 & 32 & \textbf{1.910} (0.001) & \textbf{1.624} & \textbf{91.83} & \textbf{87.42} & 0.22 & 2360448 & $1.913 \cdot 10^9$ \\
\hline
\end{tabular}
}
\end{center}
\end{table}

In the middle two experiments of table \ref{tab:t5_3}, we examine hybrids of multi-head attention and talking-heads attention, where there is a projection on one but not both of the logits and the weights.

\subsubsection{Encoder vs. Decoder}
\label{sec:encdec}

\begin{table}[h]
\caption{In each of the middle three experiments, talking-heads attention is employed in only one of the three types of attention layers in the model.}
\label{tab:t5_4}
\begin{center}
\vspace{-2mm}
\scalebox{0.75}{
\begin{tabular}{cccccc|cc|cc|ccc}
\hline\rule{0pt}{2.0ex}
     &       &     &       &       &       & ln(PPL) & ln(PPL) & SQUAD  &  & step & parameters & multiplies \\
     &       &     &       &       &       & 65536   & 524288  & v1.1   & MNLI-m & time & per  & per att. layer\\
     & $h_k$ & $h$ & $h_v$ & $d_k$ & $d_v$ & steps   & steps   & dev-f1 & dev    & (s) & att. layer  & (n=m=512) \\
\hline
multi-head &  & 24 &  & 32 & 32 & 1.989 (0.009) & 1.669 & 91.04 & 86.41 & 0.17 & 2359296 & $1.611 \cdot 10^9$ \\
TH-enc-self & 24* & 24 & 24* & 32 & 32 & 1.969 (0.002) & 1.655 & 91.63 & 87.00 & 0.21 & various & various \\
TH-dec-self & 24* & 24 & 24* & 32 & 32 & 1.981 (0.005) & 1.671 & 90.56 & 85.56 & 0.17 & various & various \\
TH-encdec & 24* & 24 & 24* & 32 & 32 & 1.942 (0.003) & 1.646 & 90.86 & 86.07 & 0.18 & various & various \\
talking-heads & 24 & 24 & 24 & 32 & 32 & \textbf{1.910} (0.001) & \textbf{1.624} & \textbf{91.83} & \textbf{87.42} & 0.22 & 2360448 & $1.913 \cdot 10^9$ \\
\hline
\end{tabular}
}
\end{center}
\end{table}

The transformer model contains three types of attention layers - self-attention in the encoder, self-attention in the decoder, and attention-over-encoder in the decoder.  In each of the middle three experiments of table \ref{tab:t5_4}, we employ talking-heads attention in only one of these types of attention layers, and multi-head attention in the others.  We find that modifying the encoder-self-attention layers has the biggest effect on the downstream language-understanding tasks.  This is unsurprising, given that these tasks have more to do with analyzing the input than with generating output.

\subsection{ALBERT}
\citep{lan2019albert} introduce ALBERT, a variation on BERT \citep{devlin2018bert}. The main difference between the ALBERT and BERT architectures is that ALBERT shares layer parameters among all layers, significantly reducing the number of parameters. For example, a 12-layer ALBERT model has about 1/12 the number of parameters in the attention and feed-forward layers as a similar BERT model.  Another difference is that the ALBERT model factorizes the word embedding as the product of two matrices with smaller bases, again significantly reducing the parameter count.  This makes ALBERT appealing for memory limited devices such as mobiles. Besides above architecture differences, ALBERT also uses sentence order prediction (SOP) to replace next sentence prediction (NSP) in BERT.

We report here experiments done with the base setting for ALBERT: a Transformer network with 12 layers of attention, the hidden and embedding size set to 768. The pre-training and fine-tuning hyperparameter settings are also exactly the same as in \citep{lan2019albert}. We use the English Wikipedia and book corpus datasets \citep{devlin2018bert}  to pre-train various models with different head sizes and talking-heads configurations. We evaluate the resulting representations by using them as a starting point to finetune for the SQuAD task (SQuAD1.1, SQuAD2.0 dev set) and various tasks (MNLI, SST-2, RACE) from the GLUE benchmark. Results are in Table \ref{tab:albert}.


\begin{table}[h]
\caption{Multi-Head vs. Talking-Heads attention on ALBERT. }
\label{tab:albert}
\begin{center}
\vspace{-2mm}
\scalebox{0.75}{
\begin{tabular}{ccc|ccccc|ccc}
\hline\rule{0pt}{2.0ex}
& heads &  $d_k = d_v$ & SQuAD1.1 (f1) &  SQuAD2.0 (f1) &  MNLI   &   SST-2 &  RACE & MLM & SOP &  Average \\ \hline
Multi-head &  6  & 128 &    88.5   &     78.8   &   79.9  &  88.6    & 62.7  &  54.3  &  85.9    & 79.7 \\
Multi-head & 12  & 64 &     88.8   &     79.3   &   80.2  &  89.9    & 63.4  &  54.5  &  86.2    & 80.32 \\
Multi-head & 24  & 32 &     88.8   &     79.1   &   79.9  &  87.7    & 62.1  &  54.4  &  85.9    & 79.52 \\
Multi-head & 48  & 16 &     87.9   &     78.8   &   79.6  &  88.4    & 61.8  &  53.8  &  85.3    & 79.3 \\ \hline
Talking-heads  &  6  & 128 &     88.7   &     78     &   80    &  88.5    & 62    &  54.1  &  85.2    & 79.44 \\
Talking-heads  & 12  & 64 &     89.2   &     79.9   &   80.5  &  89      & 65.3  &  54.9  &  \textbf{87.6}    & 80.78 \\
Talking-heads  & 24  & 32 &     89.3   &     80.5   &   80.5  &  87.6    & 65.6  &  55.3  &  86.3    & 80.7 \\
Talking-heads  & 48  & 16 &    \textbf{89.6}   &     \textbf{80.9}   &   \textbf{80.9}  &  \textbf{89.3}    & \textbf{66.5}  &  \textbf{55.7}  &  86.5    & \textbf{81.44} \\
\hline
\end{tabular}
}
\end{center}
\end{table}

We find that as the number of heads increases beyond 12 and the dimensionality of the attention-keys and attention-values decreases below 64, the performance of multi-head attention decays. On the other hand, the performance of talking-head attention keeps improving.

In addition, we also compare the logits projection and the weight projection separately with multi-head and talking-heads attention. The results are shown in Table \ref{tab:albert-logit-weight-only}. Similar to our observation in T5 experiments, only applying either the logits projection or the weight projection does not result in significant improvement compared to without them. These results again confirm the importance of having both projections.    
\begin{table}[h]
\caption{The logits-projection and the weights-projection can be employed separately.}
\label{tab:albert-logit-weight-only}
\begin{center}
\vspace{-2mm}
\scalebox{0.75}{
\begin{tabular}{ccc|ccccc|ccc}
\hline\rule{0pt}{2.0ex}
& heads & $d_k = d_v$ & SQuAD1.1 (f1) &  SQuAD2.0 (f1) &  MNLI   &   SST-2 &  RACE & MLM & SOP &  Average \\ \hline
Multi-head & 12   & 64  &     88.8   &     79.3   &   80.2  &  \textbf{89.9}    & 63.4  &  54.5  &  86.2    & 80.32 \\
Logit-project-only  & 12  & 64  &  88.5 &  78.8  &  79.8  &  89.3 &   63  &    54.6  & 85.8   & 79.88 \\
Weight-project-only & 12   & 64 &  88.9 & 79.6   &   80.3 & 89  &    64  &    54.7 &   85.8  &  80.36 \\
Talking-heads  & 12   & 64  &     \textbf{89.2}   &     \textbf{79.9}   &   \textbf{80.5}  &  89      & \textbf{65.3}  &  \textbf{54.9}  &  \textbf{87.6}    & \textbf{80.78} \\
\hline
\end{tabular}
}
\end{center}
\end{table}

\subsection{BERT}
We test various configurations of talking-heads attention based on \citep{devlin2018bert}. All of our experiments use the simplified relative position embeddings \citep{raffel2019exploring} instead of fixed position embedding. We first pre-train a 12 Transformer layers using the same dataset as \citep{devlin2018bert}. And then we finetune for the SQuAD1.1 task and the MNLI from the GLUE dataset.  Our experiments show that quality continues to improve when we grow the number of heads up to 768 and decrease the key and value dimensionality down to 1 \footnote{These extreme hyperparameter settings likely have no practical use, due to the massive amount of computation.} 

\begin{table}[h]
\caption{Talking-Heads attention on BERT. }
\label{tab:bert}
\begin{center}
\vspace{-2mm}
\scalebox{0.75}{
\begin{tabular}{ccc|ccccc|ccc}
\hline\rule{0pt}{2.0ex}
& heads  &  $d_k = d_v$&   SQuAD1.1 (f1) & MNLI   \\ \hline
Multi-head & 12 & 64 &     88.51  &     82.6 \\ \hline
Talking-heads  &  6  & 128 &     88.8   &     83.4 \\
Talking-heads   & 12  & 64  &     89.2   &     83.6 \\
Talking-heads   & 24  & 32 &     89.4   &     83.6 \\
Talking-heads   & 48  & 16 &     89.5   &     83.4 \\
Talking-heads   & 64  & 12 &     89.9   &     83.8 \\
Talking-heads   & 96  & 8 &     89.3   &     83.6 \\
Talking-heads   & 192  & 4 &    89.8   &     83.9 \\
Talking-heads   & 384  & 2 &     \textbf{90.5}   &     83.9 \\
Talking-heads   & 768  & 1 &     \textbf{90.5}   &     \textbf{84.2} \\
\hline
\end{tabular}
}
\end{center}
\end{table}

\subsection{Visualizing the Projection Matrices of Talking-Heads}
\begin{figure}
\begin{center}
   \includegraphics[width=1.0\linewidth]{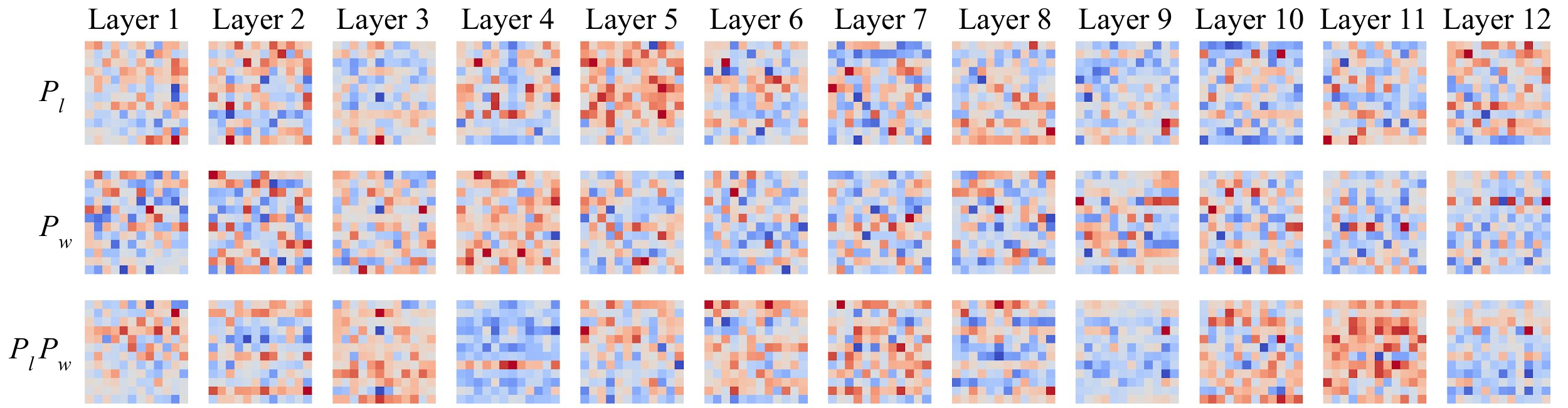}
\end{center}
   \caption{Visualization of the learned projection matrices $P_l$, $P_w$, and the multiplication $P_l \times P_w$. All entries in each matrix are normalized to [-1, 1]. The fact that these matrices are not diagonal or sparse at all, shows that there are significant data exchange across different attention heads.}
\label{fig:vis_talking_heads}
\end{figure}

To illustrate how different heads exchange information with each other, we visualize the projection matrices ($P_l$ and $P_w$) of a 12 layer BERT with 12 talking-heads in figure \ref{fig:vis_talking_heads}. Since $P_w$ is applied after $P_l$ (although there is a softmax non-linearity in between), we also visualize the combined transformation $P_l \times P_w$ in figure \ref{fig:vis_talking_heads}. As can be observed, the main diagonals of the projection matrices do not have significant greater values than other entries. This is expected because with talking-heads, a pair of query and key do not corresponds to any specific value-vector. All keys and queries jointly decide how the values in each head interchange data. Additionally, all projection matrices are well conditioned (magnitude of determinant above $10^{-9}$ with smallest eigenvalue above $10^{-3}$), indicating that no significant approximation can be achieved.

\section{Conclusions and Future Work}
We have proposed talking-heads attention and shown some promising results. One potential challenge is speed on modern deep-learning accelerators, which are optimized for large-dimension matrix multiplications.  We imagine that this will be an area of future work.  One approach is to build hardware which is better at small-dimension matrix-multiplication.  Another potential approach is to decrease the number of memory-positions considered for each query-position - for example, by using the local-attention and memory-compressed-attention approaches described in \citep{liu2018generatin}.  We look forward to more applications of talking-heads attention, as well as to further architectural improvements.

\bibliography{main}

\begin{thebibliography}{10}
\providecommand{\natexlab}[1]{#1}
\providecommand{\url}[1]{\texttt{#1}}
\expandafter\ifx\csname urlstyle\endcsname\relax
  \providecommand{\doi}[1]{doi: #1}\else
  \providecommand{\doi}{doi: \begingroup \urlstyle{rm}\Url}\fi

\bibitem[Bahdanau et~al.(2014)Bahdanau, Cho, and Bengio]{1409.0473}
Dzmitry Bahdanau, Kyunghyun Cho, and Yoshua Bengio.
\newblock Neural machine translation by jointly learning to align and
  translate, 2014.

\bibitem[Devlin et~al.(2018)Devlin, Chang, Lee, and Toutanova]{devlin2018bert}
Jacob Devlin, Ming-Wei Chang, Kenton Lee, and Kristina Toutanova.
\newblock Bert: Pre-training of deep bidirectional transformers for language
  understanding.
\newblock \emph{arXiv preprint arXiv:1810.04805}, 2018.

\bibitem[Lan et~al.(2019)Lan, Chen, Goodman, Gimpel, Sharma, and
  Soricut]{lan2019albert}
Zhenzhong Lan, Mingda Chen, Sebastian Goodman, Kevin Gimpel, Piyush Sharma, and
  Radu Soricut.
\newblock Albert: A lite bert for self-supervised learning of language
  representations, 2019.

\bibitem[Liu et~al.(2018)Liu, Saleh, Pot, Goodrich, Sepassi, Kaiser, and
  Shazeer]{liu2018generatin}
Peter~J Liu, Mohammad Saleh, Etienne Pot, Ben Goodrich, Ryan Sepassi, Lukasz
  Kaiser, and Noam Shazeer.
\newblock Generating wikipedia by summarizing long sequences.
\newblock In \emph{Proceedings of the International Conference on Learning
  Representations}, 2018.

\bibitem[Raffel et~al.(2019)Raffel, Shazeer, Roberts, Lee, Narang, Matena,
  Zhou, Li, and Liu]{raffel2019exploring}
Colin Raffel, Noam Shazeer, Adam Roberts, Katherine Lee, Sharan Narang, Michael
  Matena, Yanqi Zhou, Wei Li, and Peter Liu.
\newblock Exploring the limits of transfer learning with a unified text-to-text
  transformer.
\newblock \emph{arXiv e-prints}, 2019.

\bibitem[Rajpurkar et~al.(2016)Rajpurkar, Zhang, Lopyrev, and
  Liang]{rajpurkar2016squad}
Pranav Rajpurkar, Jian Zhang, Konstantin Lopyrev, and Percy Liang.
\newblock Squad: 100,000+ questions for machine comprehension of text.
\newblock \emph{arXiv preprint arXiv:1606.05250}, 2016.

\bibitem[Shazeer and Stern(2018)]{shazeer2018adafactor}
Noam Shazeer and Mitchell Stern.
\newblock Adafactor: Adaptive learning rates with sublinear memory cost.
\newblock \emph{arXiv preprint arXiv:1804.04235}, 2018.

\bibitem[Vaswani et~al.(2017)Vaswani, Shazeer, Parmar, Uszkoreit, Jones, Gomez,
  Kaiser, and Polosukhin]{Vas17}
Ashish Vaswani, Noam Shazeer, Niki Parmar, Jakob Uszkoreit, Llion Jones,
  Aidan~N. Gomez, Lukasz Kaiser, and Illia Polosukhin.
\newblock Attention is all you need.
\newblock In \emph{NIPS}, 2017.

\bibitem[Wang et~al.(2018)Wang, Singh, Michael, Hill, Levy, and
  Bowman]{wang2018glue}
Alex Wang, Amapreet Singh, Julian Michael, Felix Hill, Omer Levy, and Samuel~R.
  Bowman.
\newblock {GLUE}: A multi-task benchmark and analysis platform for natural
  language understanding.
\newblock \emph{arXiv preprint arXiv:1804.07461}, 2018.

\bibitem[Wang et~al.(2019)Wang, Pruksachatkun, Nangia, Singh, Michael, Hill,
  Levy, and Bowman]{wang2019superglue}
Alex Wang, Yada Pruksachatkun, Nikita Nangia, Amanpreet Singh, Julian Michael,
  Felix Hill, Omer Levy, and Samuel~R. Bowman.
\newblock Superglue: A stickier benchmark for general-purpose language
  understanding systems.
\newblock \emph{arXiv preprint arXiv:1905.00537}, 2019.

\end{thebibliography}
\bibliographystyle{plainnat}
\pagebreak
\appendix

\section{Variation: Dynamic Projections}
\label{sec:dp}

In the basic talking-heads attention algorithm described in section \ref{sec:th}, the talking-heads projections are represented by two learned weight matrices $P_l[h_k, h]$ and $ P_w[h, h_v]$.  In an additional wrinkle, we can make these projections matrices themselves input-dependent, adding terms to the projection matrices that are themselves learned linear projections of the inputs $X$ and $M$.  The algorithm is described by the pseudo-code below. \\
\begin{minipage}{\linewidth}
\begin{lstlisting}[language=Python]
def TalkingHeadsAttentionWithDynamicProjections(
    X[n, d_X],   # n vectors with dimensionality d_X
    M[m, d_M],   # m vectors with dimensionality d_M
    P_q[d_X, d_k, h_k],  # learned linear projection to produce queries
    P_k[d_M, d_k, h_k],  # learned linear projection to produce keys
    P_v[d_M, d_v, h_v],  # learned linear projection to produce values
    P_o[d_Y, d_v, h_v],  # learned linear projection of output
    P_l[h_k, h],       # learned static talking-heads proj. on logits
    P_Xl[d_X, h_k, h], # learned projection to generate dynamic talking-heads projection
    P_Ml[d_M, h_k, h], # learned projection to generate dynamic talking-heads projection
    P_w[h, h_v],       # learned static talking-heads proj. on  weights
    P_Xw[d_X, h, h_v], # learned projection to generate dynamic talking-heads projection
    P_Mw[d_X, h, h_v]) # learned projection to generate dynamic talking-heads projection
  Q[n, d_k, h_k] = einsum(X[n, d_X], P_q[d_X, d_k, h_k])  # queries         n*d_X*d_k*h_k
  K[m, d_k, h_k] = einsum(M[m, d_M], P_k[d_M, d_k, h_k])  # keys            m*d_M*d_k*h_k
  V[m, d_v, h_v] = einsum(M[m, d_M], P_v[d_M, d_v, h_v])  # values          m*d_M*d_v*h_v
  J[n, m, h_k] = einsum(Q[n, d_k, h_k], K[m, d_k, h_k])   # dot prod.       n*m*d_k*h_k
  R_Xl[n, h_k, h] = einsum(X[n, d_X], P_Xl[d_X, h_k, h])  # dynamic proj.   n*d_X*h_k*h
  R_Ml[n, h_k, h] = einsum(M[m, d_M], P_Ml[d_M, h_k, h])  # dynamic proj.   n*d_M*h_k*h
  L[n, m, h] = (
    einsum(J[n, m, h_k], P_l[h_k, h]) +       # Static talking-heads proj.  n*m*h*h_k
    einsum(J[n, m, h_k], R_Xl[n, h_k, h]) +   # Dynamic talking-heads proj. n*m*h*h_k
    einsum(J[n, m, h_k], R_Ml[m, h_k, h]))    # Dynamic talking-heads proj. n*m*h*h_k
  W[n, m, h] = softmax(L[n, m, h], reduced_dim=m)   # Attention weights
  R_Xw[n, h, h_v] = einsum(X[n, d_X], P_Xw[d_X, h, h_v])  # dynamic proj.   n*d_X*h*h_v
  R_Mw[n, h, h_v] = einsum(M[m, d_M], P_Mw[d_M, h, h_v])  # dynamic proj.   n*d_M*h*h_v
  U[n, m, h_v] = (
    einsum(W[n, m, h], P_w[h, h_v]) +         # Static Talking-heads proj.  n*m*h*h_v
    einsum(W[n, m, h], R_Xw[n, h, h_v]) +     # Dynamic talking-heads proj. n*m*h*h_v
    einsum(W[n, m, h], R_Mw[m, h, h_v]))      # Dynamic talking-heads proj. n*m*h*h_v
  O[n, d_v, h_v] = einsum(U[n, m, h_v], V[m, d_v, h_v])   #                 n*m*d_v*h_v
  Y[n, d_Y] = einsum(O[n, d_v, h_v], P_o[d_Y, d_v, h_v])  #                 n*d_Y*d_v*h_v
  return Y[n, d_Y]
\end{lstlisting}
\end{minipage}

We observed that the model only trained well if we initialized the projection-generating parameter matrices ($P_{xl}$, $P_{ml}$, $P_{xw}$, $P_{xw}$) to contain small enough values.  We used normal initializers with standard deviations of $0.1/\sqrt{d_X \cdot h_k}$, $0.1/\sqrt{d_M \cdot h_k}$, $0.1/\sqrt{d_X \cdot h}$, and $0.1/\sqrt{d_M \cdot h}$, respectively.

\begin{table}[h]
\caption{Dynamic-Projections Results on T5}
\label{tab:dp}
\begin{center}
\vspace{-2mm}
\scalebox{0.75}{
\begin{tabular}{cccccc|cc|cc|ccc}
\hline\rule{0pt}{2.0ex}
     &       &     &       &       &       & ln(PPL) & ln(PPL) & SQUAD  &  & step & parameters & multiplies \\
     &       &     &       &       &       & 65536   & 524288  & v1.1   & MNLI-m & time & per  & per att. layer\\
     & $h_k$ & $h$ & $h_v$ & $d_k$ & $d_v$ & steps   & steps   & dev-f1 & dev    & (s) & att. layer  & (n=m=512) \\
\hline
multi-head &  & 12 &  & 64 & 64 & 1.982 (0.003) & 1.678 & 90.87 & 86.20 & 0.15 & 2359296 & $1.611 \cdot 10^9$ \\
talking-heads & 12 & 12 & 12 & 64 & 64 & 1.932 (0.004) & 1.641 & 91.38 & 86.19 & 0.18 & 2359584 & $1.686 \cdot 10^9$ \\
dyn. proj. & 12 & 12 & 12 & 64 & 64 & 1.897 (0.007) & 1.595 & 90.17 & 86.18 & 0.36 & 2801952 & $1.913 \cdot 10^9$ \\
\hline
multi-head &  & 24 &  & 32 & 32 & 1.989 (0.009) & 1.669 & 91.04 & 86.41 & 0.17 & 2359296 & $1.611 \cdot 10^9$ \\
talking-heads & 24 & 24 & 24 & 32 & 32 & 1.910 (0.001) & 1.624 & \textbf{91.83} & \textbf{87.42} & 0.22 & 2360448 & $1.913 \cdot 10^9$ \\
dynamic proj. & 24 & 24 & 24 & 32 & 32 & \textbf{1.873} (0.008) & \textbf{1.587} & 90.17 & 85.94 & 0.53 & 4129920 & $2.819 \cdot 10^9$ \\
\hline
\end{tabular}
}
\end{center}
\end{table}

\subsection{Experiments}

We evaluate talking-heads attention with dynamic projections on T5 \citep{raffel2019exploring} in a set of experiments similar to those described in section \ref{sec:t5}.  

Table \ref{tab:dp} compares multi-head attention, taking-heads attention with static projections, and talking-heads attention with dynamic projections.  The dynamic projections reduce perplexity on the pre-training task.  However, in our experiments, we did not see an improvement on the downstream tasks.

\begin{table}[h]
\caption{In each of the middle four experiments, only one of the dynamic projections is employed.}
\label{tab:dp_individual}
\begin{center}
\vspace{-2mm}
\scalebox{0.75}{
\begin{tabular}{cccccc|cc|cc|ccc}
\hline\rule{0pt}{2.0ex}
     &       &     &       &       &       & ln(PPL) & ln(PPL) & SQUAD  &  & step & parameters & multiplies \\
     &       &     &       &       &       & 65536   & 524288  & v1.1   & MNLI-m & time & per  & per att. layer\\
     & $h_k$ & $h$ & $h_v$ & $d_k$ & $d_v$ & steps   & steps   & dev-f1 & dev    & (s) & att. layer  & (n=m=512) \\
\hline
talking-heads & 12 & 12 & 12 & 64 & 64 & 1.932 (0.004) & 1.641 & \textbf{91.38} & 86.19 & 0.18 & 2359584 & $1.686 \cdot 10^9$ \\
dyn. proj. $P_{Xl}$ & 12 & 12 & 12 & 64 & 64 & 1.932 (0.011) & 1.634 & 91.34 & 86.32 & 0.19 & 2470176 & $1.743 \cdot 10^9$ \\
dyn. proj. $P_{Xw}$ & 12 & 12 & 12 & 64 & 64 & 1.914 (0.005) & 1.619 & 90.70 & 86.43 & 0.19 & 2470176 & $1.743 \cdot 10^9$ \\
dyn. proj. $P_{Ml}$ & 12 & 12 & 12 & 64 & 64 & 1.930 (0.010) & 1.624 & 91.14 & \textbf{86.63} & 0.24 & 2470176 & $1.743 \cdot 10^9$ \\
dyn. proj. $P_{Mw}$ & 12 & 12 & 12 & 64 & 64 & 1.917 (0.003) & 1.624 & 90.54 & 86.45 & 0.25 & 2470176 & $1.743 \cdot 10^9$ \\
dyn. proj. & 12 & 12 & 12 & 64 & 64 & \textbf{1.897} (0.007) & \textbf{1.595} & 90.17 & 86.18 & 0.36 & 2801952 & $1.913 \cdot 10^9$ \\
\hline
\end{tabular}
}
\end{center}
\end{table}

\subsubsection{Comparing the Four Dynamic Projections}

Table \ref{tab:dp_individual} examines the effects of the four dynamic projections employed individually.   The middle four rows represent experiments where only one of the four dynamic projections were employed.   These are compared to static projections (top row) and all four dynamic projections together (bottom row).

\begin{table}[h]
\caption{Effects of applying talking-heads attention (with or without dynamic projections) in the encoder only.}
\label{tab:dp_enc_only}
\begin{center}
\vspace{-2mm}
\scalebox{0.75}{
\begin{tabular}{cccccc|cc|cc|ccc}
\hline\rule{0pt}{2.0ex}
     &       &     &       &       &       & ln(PPL) & ln(PPL) & SQUAD  &  & step & parameters & multiplies \\
     &       &     &       &       &       & 65536   & 524288  & v1.1   & MNLI-m & time & per  & per att. layer\\
     & $h_k$ & $h$ & $h_v$ & $d_k$ & $d_v$ & steps   & steps   & dev-f1 & dev    & (s) & att. layer  & (n=m=512) \\
\hline
multi-head &  & 24 &  & 32 & 32 & 1.989 (0.009) & 1.669 & 91.04 & 86.41 & 0.17 & 2359296 & $1.611 \cdot 10^9$ \\
TH-enc-self & 24* & 24 & 24* & 32 & 32 & 1.969 (0.002) & 1.655 & 91.63 & \textbf{87.00} & 0.21 & various & various \\
DP-enc-self & 24* & 24 & 24* & 32 & 32 & \textbf{1.953} (0.006) & \textbf{1.639} & \textbf{91.99} & 86.97 & 0.42 & various & various \\
\hline
\end{tabular}
}
\end{center}
\end{table}

\subsubsection{Talking-Heads in Encoder Only}

In section \ref{sec:encdec} we saw that talking heads were particularly useful in the encoder part of the model.  Table \ref{tab:dp_enc_only} presents a set of experiments where the decoder uses only multi-head attention, while the encoder uses either multi-head attention (top row), talking-heads attention with static projections (middle row), or talking-heads attention with dynamic projections (bottom row).  We observe that in this case, the dynamic projections do not appear to degrade performance on the downstream tasks.

\section{T5 Fine-Tuning Full Results}
\label{sec:finetune}
Tables \ref{tab:glue}, \ref{tab:superglue} and \ref{tab:squad} present the results of fine-tuning the models in section \ref{sec:t5} and appendix \ref{sec:dp} on the GLUE \citep{wang2018glue} and SuperGlue \citep{wang2019superglue}, and  Stanford Question-Answering Dataset (SQuAD) \citep{rajpurkar2016squad} benchmarks.

\begin{table}[h]
\caption{T5 on GLUE Language-Understanding Benchmark \citep{wang2018glue} (dev).  Experiments described in Section \ref{sec:t5} and appendix \ref{sec:dp}.}
\label{tab:glue}
\begin{center}
\vspace{-2mm}
\scalebox{0.6}{
\begin{tabular}{lccccc|c|cccccccccccc}
 & & & & & & Score & CoLA & SST-2 & MRPC & MRPC & STSB & STSB & QQP & QQP & MNLIm & MNLImm & QNLI & RTE \\
     & $h_k$ & $h$ & $h_v$ & $d_k$ & $d_v$ & Average & MCC & Acc &  F1 & Acc & PCC & SCC & F1 & Acc & Acc & Acc & Acc & Acc \\
 \hline
multi-head &  & 6 &  & 128 & 128 &  $83.20$ &  $50.96$ &  $93.35$ &  $91.49$ &  $88.24$ &  $89.51$ &  $89.41$ &  $88.79$ &  $91.69$ &  $85.12$ &  $85.70$ &  $92.35$ &  $80.14$ \\
multi-head &  & 12 &  & 64 & 64 &  $84.36$ &  $55.21$ &  $93.58$ &  $92.55$ &  $89.71$ &  $90.01$ &  $89.81$ &  $88.93$ &  $91.69$ &  $86.04$ &  $86.46$ &  $92.88$ &  $81.59$ \\
multi-head &  & 24 &  & 32 & 32 &  $84.37$ &  $55.24$ &  $93.92$ &  \textbf{93.59} &  \textbf{91.18} &  $89.79$ &  $89.68$ &  $88.96$ &  $91.68$ &  $86.06$ &  $86.16$ &  $92.73$ &  $80.87$ \\
multi-head &  & 48 &  & 16 & 16 &  $84.36$ &  $55.82$ &  $93.23$ &  \textbf{93.59} &  \textbf{91.18} &  $88.99$ &  $88.88$ &  $88.74$ &  $91.63$ &  $85.17$ &  $85.86$ &  $92.29$ &  $81.59$ \\
\hline
talking-heads & 6 & 6 & 6 & 128 & 128 &  $83.64$ &  $51.53$ &  $93.69$ &  $92.17$ &  $89.22$ &  $89.30$ &  $89.18$ &  $88.69$ &  $91.54$ &  $85.92$ &  $86.68$ &  $92.77$ &  $80.87$ \\
talking-heads & 12 & 12 & 12 & 64 & 64 &  $84.42$ &  $55.22$ &  $94.38$ &  $92.50$ &  $89.46$ &  $90.71$ &  $90.49$ &  $88.99$ &  $91.77$ &  $86.11$ &  $86.37$ &  $92.93$ &  $79.78$ \\
talking-heads & 24 & 24 & 24 & 32 & 32 &  $84.75$ &  $55.39$ &  $93.92$ &  $92.34$ &  $89.46$ &  $90.14$ &  $89.87$ &  \textbf{89.19} &  $91.91$ &  \textbf{87.42} &  $87.12$ &  $93.37$ &  $82.31$ \\
talking-heads & 48 & 48 & 48 & 16 & 16 &  $84.82$ &  $52.08$ &  \textbf{94.61} &  $92.97$ &  $90.44$ &  \textbf{91.16} &  \textbf{91.00} &  $88.95$ &  $91.78$ &  $87.40$ &  $87.44$ &  $93.32$ &  $83.03$ \\
\hline
multi-head &  & 24 &  & 64 & 64 &  $84.82$ &  \textbf{55.99} &  $94.04$ &  $92.45$ &  $89.71$ &  $90.25$ &  $90.00$ &  $89.15$ &  $91.90$ &  $86.54$ &  $86.62$ &  $93.04$ &  $81.23$ \\
general bilinear &  & 12 &  & 768 & 768 &  $84.62$ &  $53.47$ &  $93.92$ &  $93.12$ &  $90.44$ &  $90.10$ &  $89.84$ &  $89.19$ &  \textbf{91.98} &  $86.14$ &  $86.45$ &  $93.25$ &  $82.31$ \\
\hline
talking-heads & 6 & 6 & 6 & 128 & 128 &  $83.64$ &  $51.53$ &  $93.69$ &  $92.17$ &  $89.22$ &  $89.30$ &  $89.18$ &  $88.69$ &  $91.54$ &  $85.92$ &  $86.68$ &  $92.77$ &  $80.87$ \\
talking-heads & 6 & 24 & 6 & 128 & 128 &  $84.08$ &  $50.87$ &  $94.27$ &  $93.10$ &  $90.44$ &  $89.76$ &  $89.49$ &  $88.75$ &  $91.63$ &  $86.29$ &  $86.32$ &  $92.81$ &  $82.67$ \\
talking-heads & 24 & 6 & 24 & 32 & 32 &  $83.71$ &  $51.11$ &  $93.81$ &  $91.59$ &  $88.48$ &  $90.03$ &  $89.88$ &  $89.02$ &  $91.78$ &  $86.08$ &  $86.63$ &  $92.97$ &  $79.42$ \\
talking-heads & 6 & 24 & 24 & 128 & 32 &  $84.18$ &  $53.66$ &  $93.92$ &  $93.07$ &  $90.44$ &  $89.82$ &  $89.80$ &  $89.04$ &  $91.84$ &  $86.31$ &  $86.48$ &  $92.82$ &  $81.59$ \\
talking-heads & 24 & 24 & 6 & 32 & 128 &  \textbf{84.85} &  $51.80$ &  $94.04$ &  $93.12$ &  $90.44$ &  $89.90$ &  $89.75$ &  $89.15$ &  $91.83$ &  $86.81$ &  $86.81$ &  $93.08$ &  \textbf{84.84} \\
talking-heads & 24 & 24 & 24 & 32 & 32 &  $84.75$ &  $55.39$ &  $93.92$ &  $92.34$ &  $89.46$ &  $90.14$ &  $89.87$ &  \textbf{89.19} &  $91.91$ &  \textbf{87.42} &  $87.12$ &  $93.37$ &  $82.31$ \\
\hline
multi-head &  & 24 &  & 32 & 32 &  $84.37$ &  $55.24$ &  $93.92$ &  \textbf{93.59} &  \textbf{91.18} &  $89.79$ &  $89.68$ &  $88.96$ &  $91.68$ &  $86.06$ &  $86.16$ &  $92.73$ &  $80.87$ \\
project logits & 24 & 24 &  & 32 & 32 &  $84.24$ &  $54.08$ &  $93.92$ &  $91.81$ &  $88.73$ &  $90.38$ &  $90.25$ &  $89.03$ &  $91.78$ &  $85.86$ &  $86.30$ &  $92.92$ &  $81.59$ \\
project weights &  & 24 & 24 & 32 & 32 &  $83.95$ &  $51.28$ &  $93.92$ &  $92.23$ &  $89.22$ &  $89.69$ &  $89.47$ &  $88.96$ &  $91.77$ &  $86.04$ &  $86.13$ &  $93.10$ &  $82.67$ \\
talking-heads & 24 & 24 & 24 & 32 & 32 &  $84.75$ &  $55.39$ &  $93.92$ &  $92.34$ &  $89.46$ &  $90.14$ &  $89.87$ &  \textbf{89.19} &  $91.91$ &  \textbf{87.42} &  $87.12$ &  $93.37$ &  $82.31$ \\
\hline
multi-head &  & 24 &  & 32 & 32 &  $84.37$ &  $55.24$ &  $93.92$ &  \textbf{93.59} &  \textbf{91.18} &  $89.79$ &  $89.68$ &  $88.96$ &  $91.68$ &  $86.06$ &  $86.16$ &  $92.73$ &  $80.87$ \\
TH-enc-self & 24* & 24 & 24* & 32 & 32 &  $84.49$ &  $52.80$ &  $94.38$ &  $93.07$ &  $90.44$ &  $89.70$ &  $89.61$ &  $88.99$ &  $91.79$ &  $87.00$ &  $86.80$ &  $93.12$ &  $83.39$ \\
TH-dec-self & 24* & 24 & 24* & 32 & 32 &  $84.22$ &  $55.22$ &  $94.15$ &  $92.61$ &  $89.71$ &  $89.92$ &  $89.85$ &  $88.83$ &  $91.71$ &  $85.56$ &  $86.06$ &  $92.79$ &  $81.23$ \\
TH-encdec & 24* & 24 & 24* & 32 & 32 &  $84.32$ &  $54.58$ &  $93.92$ &  $92.17$ &  $88.97$ &  $89.76$ &  $89.67$ &  $89.00$ &  $91.78$ &  $86.07$ &  $85.82$ &  $92.99$ &  $80.51$ \\
talking-heads & 24 & 24 & 24 & 32 & 32 &  $84.75$ &  $55.39$ &  $93.92$ &  $92.34$ &  $89.46$ &  $90.14$ &  $89.87$ &  \textbf{89.19} &  $91.91$ &  \textbf{87.42} &  $87.12$ &  $93.37$ &  $82.31$ \\
\hline
multi-head &  & 12 &  & 64 & 64 &  $84.36$ &  $55.21$ &  $93.58$ &  $92.55$ &  $89.71$ &  $90.01$ &  $89.81$ &  $88.93$ &  $91.69$ &  $86.04$ &  $86.46$ &  $92.88$ &  $81.59$ \\
talking-heads & 12 & 12 & 12 & 64 & 64 &  $84.42$ &  $55.22$ &  $94.38$ &  $92.50$ &  $89.46$ &  $90.71$ &  $90.49$ &  $88.99$ &  $91.77$ &  $86.11$ &  $86.37$ &  $92.93$ &  $79.78$ \\
dyn. proj. & 12 & 12 & 12 & 64 & 64 &  $83.81$ &  $49.75$ &  $93.81$ &  $92.20$ &  $89.22$ &  $90.27$ &  $90.15$ &  $89.13$ &  $91.83$ &  $86.11$ &  $86.17$ &  $92.07$ &  $80.51$ \\
\hline
multi-head &  & 24 &  & 32 & 32 &  $84.37$ &  $55.24$ &  $93.92$ &  \textbf{93.59} &  \textbf{91.18} &  $89.79$ &  $89.68$ &  $88.96$ &  $91.68$ &  $86.06$ &  $86.16$ &  $92.73$ &  $80.87$ \\
talking-heads & 24 & 24 & 24 & 32 & 32 &  $84.75$ &  $55.39$ &  $93.92$ &  $92.34$ &  $89.46$ &  $90.14$ &  $89.87$ &  \textbf{89.19} &  $91.91$ &  \textbf{87.42} &  $87.12$ &  $93.37$ &  $82.31$ \\
dyn. proj. & 24 & 24 & 24 & 32 & 32 &  $83.25$ &  $49.42$ &  $93.23$ &  $92.55$ &  $89.71$ &  $89.64$ &  $89.39$ &  $88.80$ &  $91.57$ &  $85.94$ &  $85.91$ &  $92.35$ &  $79.06$ \\
\hline
talking-heads & 12 & 12 & 12 & 64 & 64 &  $84.42$ &  $55.22$ &  $94.38$ &  $92.50$ &  $89.46$ &  $90.71$ &  $90.49$ &  $88.99$ &  $91.77$ &  $86.11$ &  $86.37$ &  $92.93$ &  $79.78$ \\
dyn. proj. $P_{Xl}$ & 12 & 12 & 12 & 64 & 64 &  $84.42$ &  $53.14$ &  $94.15$ &  $91.45$ &  $87.99$ &  $90.01$ &  $89.98$ &  $88.79$ &  $91.64$ &  $86.13$ &  $86.72$ &  $93.34$ &  $81.23$ \\
dyn. proj. $P_{Xw}$ & 12 & 12 & 12 & 64 & 64 &  $84.35$ &  $55.98$ &  $94.04$ &  $93.10$ &  $90.44$ &  $89.79$ &  $89.74$ &  $88.88$ &  $91.68$ &  $86.43$ &  $86.34$ &  $92.40$ &  $80.87$ \\
dyn. proj. $P_{Ml}$ & 12 & 12 & 12 & 64 & 64 &  $84.09$ &  $51.86$ &  $94.27$ &  $92.17$ &  $89.22$ &  $89.91$ &  $89.80$ &  $89.15$ &  $91.88$ &  $86.35$ &  $87.18$ &  $92.77$ &  $80.51$ \\
dyn. proj. $P_{Mw}$ & 12 & 12 & 12 & 64 & 64 &  $84.20$ &  $54.17$ &  $93.58$ &  $92.47$ &  $89.71$ &  $89.57$ &  $89.62$ &  $88.93$ &  $91.65$ &  $86.45$ &  $86.27$ &  $92.90$ &  $81.95$ \\
dyn. proj. & 12 & 12 & 12 & 64 & 64 &  $83.81$ &  $49.75$ &  $93.81$ &  $92.20$ &  $89.22$ &  $90.27$ &  $90.15$ &  $89.13$ &  $91.83$ &  $86.11$ &  $86.17$ &  $92.07$ &  $80.51$ \\
\hline
multi-head &  & 24 &  & 32 & 32 &  $84.37$ &  $55.24$ &  $93.92$ &  \textbf{93.59} &  \textbf{91.18} &  $89.79$ &  $89.68$ &  $88.96$ &  $91.68$ &  $86.06$ &  $86.16$ &  $92.73$ &  $80.87$ \\
TH-enc-self & 24* & 24 & 24* & 32 & 32 &  $84.49$ &  $52.80$ &  $94.38$ &  $93.07$ &  $90.44$ &  $89.70$ &  $89.61$ &  $88.99$ &  $91.79$ &  $87.00$ &  $86.80$ &  $93.12$ &  $83.39$ \\
DP-enc-self & 24* & 24 & 24* & 32 & 32 &  $84.08$ &  $51.54$ &  $94.27$ &  $91.88$ &  $88.48$ &  $90.24$ &  $90.11$ &  $89.18$ &  $91.90$ &  $86.81$ &  \textbf{87.52} &  \textbf{93.48} &  $82.67$ \\
\hline
\hline
\citep{raffel2019exploring} &  & 12 &  & 64 & 64 &  $83.49$ & $53.90$ & $92.43$ & $92.25$ & $89.46$ & $87.49$ & $87.53$ & $88.72$ & $91.51$ & $84.85$ & $84.84$ & $90.99$ & $77.26$  \\
ibid. stddev. & &  &  &  &  & $0.235$ & $1.111$ & $0.569$ & $0.729$ & $1.019$ & $0.374$ & $0.418$ & $0.108$ & $0.070$ & $0.291$ & $0.231$ & $0.361$ & $1.393$ \\

\end{tabular}
}
\end{center}
\end{table}

\begin{table}[h]
\caption{T5 on SuperGLUE Language-Understanding Benchmark \citep{wang2019superglue} (dev). Experiments described in Section \ref{sec:t5} and appendix \ref{sec:dp}.}
\label{tab:superglue}
\begin{center}
\vspace{-2mm}
\scalebox{0.6}{
\begin{tabular}{lccccc|c|cccccccccccc}
 & & & & &  & Score & BoolQ & CB  & CB & CoPA & MultiRC & MultiRC & ReCoRD & ReCoRD  & RTE & WiC & WSC \\
   & $h_k$ & $h$ & $h_v$ & $d_k$ & $d_v$ & Average & Acc & F1 & Acc & Acc & F1 & EM & F1 & EM & Acc & Acc & Acc \\
\hline
multi-head &  & 6 &  & 128 & 128 &  $72.00$ &  $78.84$ &  $78.83$ &  $87.50$ &  $70.00$ &  $75.01$ &  $36.94$ &  $72.08$ &  $71.32$ &  $83.03$ &  $69.12$ &  $79.81$ \\
multi-head &  & 12 &  & 64 & 64 &  $73.59$ &  $80.31$ &  $89.15$ &  $89.29$ &  $73.00$ &  $75.13$ &  $37.57$ &  $73.92$ &  $72.94$ &  $83.75$ &  $69.28$ &  $77.88$ \\
multi-head &  & 24 &  & 32 & 32 &  $73.98$ &  $81.35$ &  $84.24$ &  $91.07$ &  $70.00$ &  $75.98$ &  $39.24$ &  $74.40$ &  $73.37$ &  $83.03$ &  $71.00$ &  $78.85$ \\
multi-head &  & 48 &  & 16 & 16 &  $71.85$ &  $80.34$ &  $77.90$ &  $87.50$ &  $67.00$ &  $75.26$ &  $37.67$ &  $72.31$ &  $71.32$ &  $84.12$ &  $69.44$ &  $77.88$ \\
\hline
talking-heads & 6 & 6 & 6 & 128 & 128 &  $72.57$ &  $80.83$ &  $83.37$ &  $89.29$ &  $65.00$ &  $76.76$ &  $40.08$ &  $75.37$ &  $74.48$ &  $83.39$ &  $66.30$ &  $80.77$ \\
talking-heads & 12 & 12 & 12 & 64 & 64 &  $73.16$ &  $81.38$ &  $81.13$ &  $89.29$ &  $69.00$ &  $77.36$ &  $40.50$ &  $76.45$ &  $75.63$ &  $83.39$ &  $65.83$ &  $82.69$ \\
talking-heads & 24 & 24 & 24 & 32 & 32 &  $75.80$ &  $81.96$ &  \textbf{90.60} &  \textbf{94.64} &  $75.00$ &  $76.56$ &  $39.77$ &  $77.22$ &  $76.37$ &  $85.20$ &  $70.06$ &  $81.73$ \\
talking-heads & 48 & 48 & 48 & 16 & 16 &  \textbf{76.39} &  \textbf{82.94} &  $87.46$ &  $91.07$ &  $74.00$ &  $78.11$ &  $42.71$ &  $77.51$ &  $76.68$ &  $84.84$ &  $69.59$ &  \textbf{87.50} \\
\hline
multi-head &  & 24 &  & 64 & 64 &  $74.08$ &  $81.80$ &  $77.90$ &  $87.50$ &  $73.00$ &  $77.20$ &  $39.35$ &  $76.89$ &  $76.11$ &  $83.39$ &  $69.44$ &  $80.77$ \\
general bilinear &  & 12 &  & 768 & 768 &  $73.35$ &  $81.47$ &  $83.46$ &  $89.29$ &  $71.00$ &  $76.69$ &  $39.24$ &  $76.80$ &  $76.02$ &  $85.92$ &  $69.59$ &  $76.92$ \\
\hline
talking-heads & 6 & 6 & 6 & 128 & 128 &  $72.57$ &  $80.83$ &  $83.37$ &  $89.29$ &  $65.00$ &  $76.76$ &  $40.08$ &  $75.37$ &  $74.48$ &  $83.39$ &  $66.30$ &  $80.77$ \\
talking-heads & 6 & 24 & 6 & 128 & 128 &  $73.71$ &  $81.07$ &  $80.92$ &  $87.50$ &  $68.00$ &  $76.47$ &  $40.82$ &  $74.77$ &  $73.89$ &  $85.92$ &  $70.06$ &  $79.81$ \\
talking-heads & 24 & 6 & 24 & 32 & 32 &  $73.56$ &  $80.92$ &  $83.52$ &  $89.29$ &  $75.00$ &  $75.64$ &  $37.15$ &  $74.58$ &  $73.73$ &  $81.95$ &  \textbf{71.32} &  $76.92$ \\
talking-heads & 6 & 24 & 24 & 128 & 32 &  $74.29$ &  $80.95$ &  $87.62$ &  $91.07$ &  $74.00$ &  $76.23$ &  $37.04$ &  $76.67$ &  $75.83$ &  $83.39$ &  $68.65$ &  $82.69$ \\
talking-heads & 24 & 24 & 6 & 32 & 128 &  $76.37$ &  $81.77$ &  $88.97$ &  $92.86$ &  \textbf{76.00} &  $77.63$ &  $42.81$ &  $76.72$ &  $75.88$ &  \textbf{86.28} &  $67.71$ &  \textbf{87.50} \\
talking-heads & 24 & 24 & 24 & 32 & 32 &  $75.80$ &  $81.96$ &  \textbf{90.60} &  \textbf{94.64} &  $75.00$ &  $76.56$ &  $39.77$ &  $77.22$ &  $76.37$ &  $85.20$ &  $70.06$ &  $81.73$ \\
\hline
multi-head &  & 24 &  & 32 & 32 &  $73.98$ &  $81.35$ &  $84.24$ &  $91.07$ &  $70.00$ &  $75.98$ &  $39.24$ &  $74.40$ &  $73.37$ &  $83.03$ &  $71.00$ &  $78.85$ \\
project logits & 24 & 24 &  & 32 & 32 &  $72.63$ &  $81.47$ &  $83.15$ &  $89.29$ &  $71.00$ &  $76.98$ &  $39.35$ &  $75.00$ &  $74.21$ &  $85.20$ &  $69.75$ &  $79.81$ \\
project weights &  & 24 & 24 & 32 & 32 &  $74.05$ &  $81.99$ &  $81.96$ &  $87.50$ &  $73.00$ &  $77.35$ &  $41.03$ &  $76.62$ &  $75.74$ &  $85.20$ &  $68.03$ &  $78.85$ \\
talking-heads & 24 & 24 & 24 & 32 & 32 &  $75.80$ &  $81.96$ &  \textbf{90.60} &  \textbf{94.64} &  $75.00$ &  $76.56$ &  $39.77$ &  $77.22$ &  $76.37$ &  $85.20$ &  $70.06$ &  $81.73$ \\
\hline
multi-head &  & 24 &  & 32 & 32 &  $73.98$ &  $81.35$ &  $84.24$ &  $91.07$ &  $70.00$ &  $75.98$ &  $39.24$ &  $74.40$ &  $73.37$ &  $83.03$ &  $71.00$ &  $78.85$ \\
TH-enc-self & 24* & 24 & 24* & 32 & 32 &  $73.90$ &  $82.72$ &  $84.20$ &  $89.29$ &  $69.00$ &  \textbf{78.18} &  \textbf{43.55} &  $76.07$ &  $75.34$ &  $85.92$ &  $69.75$ &  $79.81$ \\
TH-dec-self & 24* & 24 & 24* & 32 & 32 &  $72.54$ &  $80.92$ &  $79.93$ &  $87.50$ &  $67.00$ &  $76.49$ &  $38.09$ &  $73.72$ &  $72.94$ &  $83.03$ &  $69.44$ &  $79.81$ \\
TH-encdec & 24* & 24 & 24* & 32 & 32 &  $73.44$ &  $80.83$ &  $78.67$ &  $87.50$ &  $74.00$ &  $76.41$ &  $39.24$ &  $75.35$ &  $74.47$ &  $83.39$ &  $70.53$ &  $80.77$ \\
talking-heads & 24 & 24 & 24 & 32 & 32 &  $75.80$ &  $81.96$ &  \textbf{90.60} &  \textbf{94.64} &  $75.00$ &  $76.56$ &  $39.77$ &  $77.22$ &  $76.37$ &  $85.20$ &  $70.06$ &  $81.73$ \\
\hline
multi-head &  & 12 &  & 64 & 64 &  $73.59$ &  $80.31$ &  $89.15$ &  $89.29$ &  $73.00$ &  $75.13$ &  $37.57$ &  $73.92$ &  $72.94$ &  $83.75$ &  $69.28$ &  $77.88$ \\
talking-heads & 12 & 12 & 12 & 64 & 64 &  $73.16$ &  $81.38$ &  $81.13$ &  $89.29$ &  $69.00$ &  $77.36$ &  $40.50$ &  $76.45$ &  $75.63$ &  $83.39$ &  $65.83$ &  $82.69$ \\
dyn. proj. & 12 & 12 & 12 & 64 & 64 &  $71.98$ &  $80.52$ &  $86.31$ &  $91.07$ &  $66.00$ &  $76.47$ &  $37.67$ &  $73.24$ &  $72.42$ &  $82.31$ &  $67.40$ &  $75.00$ \\
\hline
multi-head &  & 24 &  & 32 & 32 &  $73.98$ &  $81.35$ &  $84.24$ &  $91.07$ &  $70.00$ &  $75.98$ &  $39.24$ &  $74.40$ &  $73.37$ &  $83.03$ &  $71.00$ &  $78.85$ \\
talking-heads & 24 & 24 & 24 & 32 & 32 &  $75.80$ &  $81.96$ &  \textbf{90.60} &  \textbf{94.64} &  $75.00$ &  $76.56$ &  $39.77$ &  $77.22$ &  $76.37$ &  $85.20$ &  $70.06$ &  $81.73$ \\
dyn. proj. & 24 & 24 & 24 & 32 & 32 &  $72.70$ &  $79.45$ &  $78.80$ &  $87.50$ &  $72.00$ &  $74.68$ &  $35.15$ &  $72.39$ &  $71.70$ &  $80.14$ &  $68.34$ &  $82.69$ \\
\hline
talking-heads & 12 & 12 & 12 & 64 & 64 &  $73.16$ &  $81.38$ &  $81.13$ &  $89.29$ &  $69.00$ &  $77.36$ &  $40.50$ &  $76.45$ &  $75.63$ &  $83.39$ &  $65.83$ &  $82.69$ \\
dyn. proj. $P_{Xl}$ & 12 & 12 & 12 & 64 & 64 &  $74.55$ &  $81.19$ &  $86.19$ &  $91.07$ &  $74.00$ &  $76.20$ &  $41.45$ &  $75.45$ &  $74.63$ &  $84.84$ &  $68.81$ &  $82.69$ \\
dyn. proj. $P_{Xw}$ & 12 & 12 & 12 & 64 & 64 &  $73.85$ &  $80.73$ &  $82.39$ &  $89.29$ &  $74.00$ &  $75.50$ &  $38.51$ &  $73.72$ &  $72.93$ &  $83.03$ &  $68.50$ &  $81.73$ \\
dyn. proj. $P_{Ml}$ & 12 & 12 & 12 & 64 & 64 &  $74.51$ &  $81.62$ &  $86.06$ &  $92.86$ &  $72.00$ &  $75.48$ &  $38.82$ &  $75.54$ &  $74.70$ &  $83.03$ &  $68.65$ &  $82.69$ \\
dyn. proj. $P_{Mw}$ & 12 & 12 & 12 & 64 & 64 &  $73.61$ &  $80.61$ &  $79.72$ &  $89.29$ &  $70.00$ &  $75.36$ &  $38.20$ &  $75.19$ &  $74.41$ &  $83.39$ &  $67.55$ &  $83.65$ \\
dyn. proj. & 12 & 12 & 12 & 64 & 64 &  $71.98$ &  $80.52$ &  $86.31$ &  $91.07$ &  $66.00$ &  $76.47$ &  $37.67$ &  $73.24$ &  $72.42$ &  $82.31$ &  $67.40$ &  $75.00$ \\
\hline
multi-head &  & 24 &  & 32 & 32 &  $73.98$ &  $81.35$ &  $84.24$ &  $91.07$ &  $70.00$ &  $75.98$ &  $39.24$ &  $74.40$ &  $73.37$ &  $83.03$ &  $71.00$ &  $78.85$ \\
TH-enc-self & 24* & 24 & 24* & 32 & 32 &  $73.90$ &  $82.72$ &  $84.20$ &  $89.29$ &  $69.00$ &  \textbf{78.18} &  \textbf{43.55} &  $76.07$ &  $75.34$ &  $85.92$ &  $69.75$ &  $79.81$ \\
DP-enc-self & 24* & 24 & 24* & 32 & 32 &  $74.65$ &  $82.60$ &  $84.76$ &  $91.07$ &  $69.00$ &  $77.32$ &  $42.29$ &  \textbf{77.88} &  \textbf{76.99} &  $84.48$ &  $71.16$ &  $79.81$ \\
\hline\hline
\citep{raffel2019exploring} &  & 12 &  & 64 & 64  & $72.53$ & $76.85$ & $94.37$ & $94.64$ & $70.00$ & $67.64$ & $28.75$ & $70.84$ & $69.90$ & $74.73$ & $67.71$ & $77.88$ \\
ibid. stddev. &  &  &  &  &  & $0.416$ & $0.365$ & $3.237$ & $2.560$ & $2.741$ & $0.716$ & $1.011$ & $0.370$ & $0.379$ & $1.228$ & $0.850$ & $2.029$ \\

\end{tabular}
}
\end{center}
\end{table}

\begin{table}[h]
\caption{T5 on SQuAD \citep{rajpurkar2016squad} v1.1 (dev). Experiments described in Section \ref{sec:t5} and appendix \ref{sec:dp}.}
\label{tab:squad}
\begin{center}
\vspace{-2mm}
\scalebox{0.75}{
\begin{tabular}{lccccc|ll}
   & $h_k$ & $h$ & $h_v$ & $d_k$ & $d_v$ & EM & F1 \\
\hline
multi-head &  & 6 &  & 128 & 128 &  $81.87$ &  $89.88$ \\
multi-head &  & 12 &  & 64 & 64 &  $83.30$ &  $90.80$ \\
multi-head &  & 24 &  & 32 & 32 &  $83.71$ &  $91.04$ \\
multi-head &  & 48 &  & 16 & 16 &  $82.62$ &  $90.31$ \\
\hline
talking-heads & 6 & 6 & 6 & 128 & 128 &  $82.47$ &  $90.51$ \\
talking-heads & 12 & 12 & 12 & 64 & 64 &  $83.67$ &  $91.38$ \\
talking-heads & 24 & 24 & 24 & 32 & 32 &  \textbf{84.66} &  $91.74$ \\
talking-heads & 48 & 48 & 48 & 16 & 16 &  $84.38$ &  $91.85$ \\
\hline
multi-head &  & 24 &  & 64 & 64 &  $84.31$ &  $91.46$ \\
general bilinear &  & 12 &  & 768 & 768 &  $83.03$ &  $90.76$ \\
\hline
talking-heads & 6 & 6 & 6 & 128 & 128 &  $82.47$ &  $90.51$ \\
talking-heads & 6 & 24 & 6 & 128 & 128 &  $83.40$ &  $90.91$ \\
talking-heads & 24 & 6 & 24 & 32 & 32 &  $83.16$ &  $90.77$ \\
talking-heads & 6 & 24 & 24 & 128 & 32 &  $83.30$ &  $91.02$ \\
talking-heads & 24 & 24 & 6 & 32 & 128 &  $83.25$ &  $90.98$ \\
talking-heads & 24 & 24 & 24 & 32 & 32 &  \textbf{84.66} &  $91.74$ \\
\hline
multi-head &  & 24 &  & 32 & 32 &  $83.71$ &  $91.04$ \\
project logits & 24 & 24 &  & 32 & 32 &  $83.36$ &  $91.29$ \\
project weights &  & 24 & 24 & 32 & 32 &  $83.26$ &  $91.03$ \\
talking-heads & 24 & 24 & 24 & 32 & 32 &  \textbf{84.66} &  $91.74$ \\
\hline
multi-head &  & 24 &  & 32 & 32 &  $83.71$ &  $91.04$ \\
TH-enc-self & 24* & 24 & 24* & 32 & 32 &  $84.39$ &  $91.59$ \\
TH-dec-self & 24* & 24 & 24* & 32 & 32 &  $83.31$ &  $90.54$ \\
TH-encdec & 24* & 24 & 24* & 32 & 32 &  $83.27$ &  $90.86$ \\
talking-heads & 24 & 24 & 24 & 32 & 32 &  \textbf{84.66} &  $91.74$ \\
\hline
multi-head &  & 12 &  & 64 & 64 &  $83.30$ &  $90.80$ \\
talking-heads & 12 & 12 & 12 & 64 & 64 &  $83.67$ &  $91.38$ \\
dyn. proj. & 12 & 12 & 12 & 64 & 64 &  $82.16$ &  $90.13$ \\
\hline
multi-head &  & 24 &  & 32 & 32 &  $83.71$ &  $91.04$ \\
talking-heads & 24 & 24 & 24 & 32 & 32 &  \textbf{84.66} &  $91.74$ \\
dyn. proj. & 24 & 24 & 24 & 32 & 32 &  $82.09$ &  $90.17$ \\
\hline
talking-heads & 12 & 12 & 12 & 64 & 64 &  $83.67$ &  $91.38$ \\
dyn. proj. $P_{Xl}$ & 12 & 12 & 12 & 64 & 64 &  $83.61$ &  $91.29$ \\
dyn. proj. $P_{Xw}$ & 12 & 12 & 12 & 64 & 64 &  $82.86$ &  $90.70$ \\
dyn. proj. $P_{Ml}$ & 12 & 12 & 12 & 64 & 64 &  $83.30$ &  $91.14$ \\
dyn. proj. $P_{Mw}$ & 12 & 12 & 12 & 64 & 64 &  $82.78$ &  $90.53$ \\
dyn. proj. & 12 & 12 & 12 & 64 & 64 &  $82.16$ &  $90.13$ \\
\hline
multi-head &  & 24 &  & 32 & 32 &  $83.71$ &  $91.04$ \\
TH-enc-self & 24* & 24 & 24* & 32 & 32 &  $84.39$ &  $91.59$ \\
DP-enc-self & 24* & 24 & 24* & 32 & 32 &  $84.64$ &  \textbf{91.99} \\
\hline\hline
\citep{raffel2019exploring} &  & 12 &  & 64 & 64 & $81.84$ & $89.66$ \\
ibid. stddev. &  & &  &  &  & $0.343$ & $0.226$ \\
\end{tabular}
}
\end{center}
\end{table}


\end{document}